# Interpretable Multimodal Framework for Human-Centered Street Assessment: Integrating Visual-Language Models for Perceptual Urban Diagnostics

Haotian Lan

**Abstract:** While objective street metrics derived from imagery or GIS have become standard in urban analytics, they remain insufficient to capture subjective perceptions essential to inclusive urban design. This study introduces a novel Multimodal Street Evaluation Framework (MSEF) that fuses a vision transformer (VisualGLM-6B) with a large language model (GPT-4), enabling interpretable dual-output assessment of streetscapes. Leveraging over 15,000 annotated street-view images from Harbin, China, we fine-tune the framework using LoRA and P-Tuning v2 for parameter-efficient adaptation. The model achieves an F1 score of 0.84 on objective features and 89.3% agreement with aggregated resident perceptions, validated across stratified socioeconomic geographies.

Beyond classification accuracy, MSEF captures context-dependent contradictions: for instance, informal commerce boosts perceived vibrancy while simultaneously reducing pedestrian comfort. It also identifies nonlinear and semantically contingent patterns — such as the divergent perceptual effects of architectural transparency across residential and commercial zones—revealing the limits of universal spatial heuristics.

By generating natural-language rationales grounded in attention mechanisms, the framework bridges sensory data with socio-affective inference, enabling transparent diagnostics aligned with SDG 11. This work offers both methodological innovation in urban perception modeling and practical utility for planning systems seeking to reconcile infrastructural precision with lived experience.

## 1.Introduction

Urban-street quality studies have long relied on objective, image or GIS-derived indicators such as roadway width, traffic density, canopy cover, pavement integrity to gauge the performance of public space [1, 2]. While such metrics underpin infrastructure audits, they fail to capture residents' lived experience; feelings of safety, comfort, or visual pleasure often diverge sharply from physical measurements [3]. This objective subjective gap produces what we term evaluation fuzziness: two blocks with identical vehicle counts may evoke opposite comfort levels because of differences in sidewalk enclosure or façade complexity [4]. Recent evidence links this mismatch to noticeable declines in street-level social interaction and enduring impacts on neighbourhood vitality [5].

Multimodal large models (MLMs) offer a new way to bridge the divide by jointly processing visual inputs and natural-language feedback through cross-modal alignment [6, 7]. Yet existing deployments remain hampered by (i) limited field validation in complex real-world streetscapes and (ii) insufficient interpretability for policy decisions [8]. To address both limitations, we propose an interpretable vision-language framework that fuses high-resolution street imagery, large-scale resident surveys and a neural architecture designed to output human-readable rationales alongside numeric scores.

Harbin, China, is an instructive testbed: its urban fabric ranges from century-old mixed-use lanes to Soviet-era super-blocks and newly densified commercial strips. Our training corpus therefore samples panoramas city-wide, while the empirical evaluation presented here concentrates on the HLJSTU academic precinct-a university district now morphing into a hybrid commercial-residential corridor. Within 872 street segments we probe a recurring SDG-11 tension: how to balance retail vitality with sidewalk comfort for pedestrians.

To address this question we introduce a dual-output framework that links a large-language model (GPT-4) with a vision-language model (VisualGLM-6B). The vision module extracts objective cues from each panorama, and the language module translates those cues into both scalar indices and concise, plain-language rationales. By coupling numeric precision with transparent explanations, the system equips planners with evidence that resonates equally with technical audits and lived experience-advancing SDG-11's mandate for inclusive, safe and resilient public space [9].

The remainder of the article is organised as follows: Section 2 reviews the relevant literature; Section 3 describes the methodology; Section 4 outlines the data-collection process; Section 5 details model validation and field deployment; Section 6 presents the results; and Section 7 offers conclusions, limitations and avenues for future research.

## 2.Literature Review

*2.1 Instrumentalizing Streetscapes: The Reign and Limits of Objective Metrics*

The past decade has witnessed a surge in the use of computational techniques to quantify urban environments, especially through street-level imagery and GIS-derived indicators. Metrics such as roadway width, tree-canopy cover, façade complexity, and pavement integrity have become staples in evaluating public space quality, undergirding audits of walkability, safety, and environmental comfort [10–12]. Enabled by convolutional neural networks (CNNs), these approaches extract high-resolution geometric and material features at scale, generating urban analytics previously inaccessible to planners [13].

Among these, the Green View Index (GVI) exemplifies how pixel-level vegetation detection from panoramas has been operationalized to estimate resident well-being. Empirical studies in cities like Beijing and Singapore suggest that higher GVI values correspond with reported satisfaction and psychological restoration, lending legitimacy to the notion that objective greenery can act as a proxy for subjective vitality [14,15]. Likewise, emerging frameworks for pedestrian space modeling — including network-based representations of crossing density, intersection complexity, and sidewalk width — have provided valuable indicators for walkability and accessibility planning [16,17].

Yet despite their utility, these instrumental approaches face critical limitations. A growing body of evidence highlights the perceptual disconnect between what is measured and what is experienced. For instance, the widely cited StreetScore project, while technically sophisticated, was shown to systematically overrate safety in crowded commercial districts— a 30-point deviation compared to human surveys—due to its overreliance on façade visibility and neglect of micro-scale discomforts such as noise, crowding, or lighting quality [18]. This gap underscores a broader epistemological issue: infrastructural presence does not guarantee psychological comfort, and measurable form often fails to capture the full spectrum of human spatial perception [19].

More broadly, attempts to universalize metrics across contexts have encountered cultural and environmental variability. For example, while commercial density may signal vitality in Tokyo's narrow alleyways, it may register as congestion or even insecurity in less familiar or more automobile-oriented urban settings [20]. Similarly, enclosure — often linked with pedestrian safety — may have opposing effects depending on local norms regarding surveillance, gendered mobility, or nighttime activity [21]. These divergences point to a fundamental paradox in objective modeling: the very features that constitute "good design" in one locale may elicit discomfort or alienation in another.

These tensions motivate the turn toward multimodal and perceptually grounded approaches. As urban research shifts toward SDG 11's human-centered imperatives — emphasizing inclusivity, safety, and subjective well-being — new tools are needed that integrate not only what is physically present, but how it is experienced. In this light, image-based metrics must be treated not as definitive evaluations, but as partial signals to be interpreted through contextual, cultural, and psychological frames. The need for such integration provides the foundation for our proposed framework, which pairs visual representation with large-scale perceptual data to bridge the objective–subjective divide in urban street assessment.

*2.2 The Subjectivity Conundrum: From Noise to Signal*

Urban planning has long grappled with the challenge of incorporating subjective perceptions—such as delight, sociability, and perceived safety—into evidence-based design. Historically dismissed as anecdotal or "noisy," these affective responses are now recognized as integral signals within urban systems. Seminal work by Jan Gehl demonstrated that intangible qualities, including perceived comfort and vitality, account for over two-thirds of the variation in street-level social interactions, even when formal spatial parameters remain constant [22]. This paradigm shift has reoriented urban analysis toward the experiential dimension of space.

Recent crowdsourced initiatives such as StreetScore and StreetScale have operationalized this perspective by collecting pairwise image comparisons to model perceived safety across thousands of urban scenes. However, their predictive accuracy remains uneven: in Baltimore, for example, modeled perceptions diverged from resident-reported safety in over 40% of neighborhoods, particularly in racially or economically heterogeneous zones [23]. These discrepancies underscore the deeply contextual nature of perception, shaped not only by visual cues but also by cultural memory, social narratives, and temporal dynamics.

Advances in deep learning have further expanded the methodological toolkit for capturing these elusive qualities. Convolutional neural networks trained on large-scale image datasets now allow researchers to quantify multi-dimensional perceptual constructs such as historic charm, complexity, or visual disorder across entire urban landscapes [24]. In Shanghai, an affective atlas derived from such techniques has begun informing municipal revitalization projects, suggesting a growing institutional appetite for perception-driven planning [25]. Meanwhile, temporally tagged social media data reveals the fluidity of affect: streets praised as vibrant and welcoming during daylight hours may be described as hostile or eerie at night, illustrating how urban perception oscillates across time and use cycles [26].

Crucially, hybrid models integrating both objective and subjective indicators offer stronger explanatory power than either domain alone. For instance, recent studies show that combining Green View Index (GVI) with perceived walkability scores significantly improves predictions of life satisfaction and mental well-being compared to vegetation coverage alone [27, 28]. These findings affirm that subjective experience is not merely a soft correlate but a structural determinant of urban quality.

Nevertheless, the path toward scalable and trustworthy perceptual modeling is fraught with challenges. Traditional survey instruments are labor-intensive, geographically limited, and often lack transparency. On the other hand, black-box models—while efficient—raise concerns about interpretability and public accountability, especially in planning contexts with high social stakes [29]. As cities increasingly seek to algorithmically mediate the lived experience of space, the central conundrum remains: how can we move from noisy perception to meaningful signal without losing nuance, agency, or trust?

*2.3 Multimodal Synergy: Bridging Sensors and Sentiments*

The integration of vision-language models (VLMs) into urban analysis presents a powerful avenue for bridging the long-standing divide between sensory data and socio-affective interpretation. These multimodal systems translate raw visual inputs into semantically rich urban descriptors, enabling machines to infer not just what is seen, but what is felt. CLIP-based frameworks, for instance, have demonstrated statistically significant correlations between visual features — such as colonnaded façades, neon signage, or wall textures — and crowd-sourced descriptors like "historic," "chaotic," or "vibrant" [30]. Such associations suggest that urban form can be computationally re-situated within subjective cultural taxonomies.

Emerging platforms operationalize this capacity. SAGAI (Street Attribute Guided AI) enables prompt-based querying ("Assess storefront vibrancy"), returning geocoded outputs suitable for planning applications. UrbanCLIP further shows that image-text co-embedding improves socio-economic inference accuracy over image-only models, especially in morphologically ambiguous streetscapes [31, 32]. These systems thus exemplify a shift from passive recognition to interactive diagnosis.

Yet, substantive barriers to adoption persist. Cultural fragility remains a pressing concern: models fine-tuned on dominant or Western-centric datasets often misclassify immigrant or informal districts, interpreting visual heterogeneity as disorder—a cross-context generalization failure observed in multiple global case studies [31]. Moreover, most models remain temporally myopic, ignoring perception volatility across the day-night cycle. Social media analyses, however, consistently reveal diurnal affective swings—where, for example, bustling boulevards are described as energetic by day and threatening by night [33, 32].

Opacity is another structural limitation. Despite recent progress in explainable AI, such as the use of Dynamic Accumulated Attention Maps to visualize token-to-region alignment [35], practitioners remain wary of opaque model outputs, particularly when deployed in socially contested spaces. Interpretability is not merely a technical add-on but a condition for institutional trust and public legitimacy.

To overcome these obstacles, two methodological directions have emerged. The first is perceptual dimension decomposition, which reduces holistic impressions (e.g., "pleasantness") into a composite of interpretable factors like cleanliness, greenness, and spatial enclosure [34]. The second involves multi-label classification frameworks using attention-based mechanisms to model the co-occurrence and interactions between perceptual dimensions—for example, how perceived safety might modulate or suppress impressions of commercial vibrancy [35].

*2.4 Why Multimodal Large Models Matter: From Description to Deliberation*

Multimodal large language models (MLLMs)—notably GPT-4V, CogVLM, and their successors—represent a paradigm shift in urban perception modeling. Unlike earlier systems that relied on convolutional backbones and pre-trained visual encoders for classification or segmentation tasks, these models unify language reasoning and visual understanding in a single architecture. Their strength lies not merely in image captioning or tagging, but in interpreting urban scenes through flexible, dialogic, and temporally grounded language.

MLLMs offer remarkable semantic adaptability. Given a street-level image, GPT-4V or CogVLM can respond to diverse natural-language prompts ranging from factual ("How many storefronts are visible?") to evaluative ("Does this street feel safe for children?") without requiring task-specific tuning. This generalization capacity enables researchers to probe latent perceptual dimensions — such as serenity, commercial vitality, or social inclusivity — even in data-scarce or culturally unfamiliar settings, making these models particularly suited to comparative urban research and adaptive design interventions [36, 37, 38].

Beyond description, MLLMs provide reasoning capacity in natural language. When asked to evaluate pedestrian comfort, for example, a model might cite narrow sidewalks, obstructed sightlines, or the absence of shade as contributing factors. These textual rationales, grounded in visual evidence, bridge the gap between technical diagnostics and stakeholder interpretation, supporting more deliberative and accountable planning practices [39, 40].

Their capacity for interactive exploration further expands their utility. In co-analysis settings, these models can simulate multiple stakeholder viewpoints or respond to iterative prompts refined by role or concern—such as re-evaluating a street from the perspective of elderly pedestrians or nighttime visitors. In this way, MLLMs transform static perception modeling into a medium for negotiated interpretation, aligned with contemporary values of procedural equity and public participation [41, 42].

Temporal sensitivity is another distinctive advantage. Unlike earlier image-based models that ignore time, MLLMs can incorporate time-stamped imagery and adapt their textual inferences based on known diurnal patterns. This opens new avenues for understanding how street atmospheres fluctuate over time, particularly when combined with insights from social media or sensor-linked datasets [43]. Moreover, interpretability tools—such as attention maps or token-region alignment—make it possible to trace the internal logic of model outputs, helping to counteract black-box skepticism in policy settings [44, 45].

In sum, GPT-4V and CogVLM do not simply enrich urban scene understanding—they reframe it as a process of narrative construction. Their ability to render streets not just as geometric objects but as culturally negotiated experiences repositions MLLMs as deliberative mediators between data, discourse, and design.

## 3. Methodology

*3.1. Model Architecture and Adaptation Workflow*

Our pipeline begins with street-view panoramas that are encoded by a ViT backbone enhanced with BLIP-2 Q-Former modules; this stage compresses geometric cues—sidewalk layout, façade articulation, tree-canopy texture—into 32 latent tokens suitable for language-model processing. These image tokens, along with coarse scene metadata, are then supplied to GPT-4, whose role is to translate residents' answers from a standardised questionnaire into a uniform appraisal and to assign provisional scalar tags for walkability, enclosure, greenery and vibrancy. The resulting soft labels both offer an interpretable ground truth for auditors and seed the fine-tuning of VisualGLM-6B, which serves as the lightweight inference "student."

To specialise VisualGLM-6B for streetscape analytics without sacrificing its general knowledge, we use a two-step adaptation scheme. First, Low-Rank Adaptation (LoRA) injects rank-8 matrices into the Q-Former and the upper transformer layers, recalibrating visual attention while leaving 98 % of the base weights frozen:

$$\mathbf{W}^{\text{LoRA}} = \mathbf{W}_0 + \underbrace{\mathbf{AB}}_{\Delta \mathbf{W}}, \quad \mathbf{A} \in \mathbb{R}^{d \times r}, \mathbf{B} \in \mathbb{R}^{r \times d}, r \ll d \tag{1}$$

where $W_0$ is the frozen pretrained weight matrix

A and B are the trainable low-rank factors, d is the layer width, and r = 8 is the chosen rank.

This LoRA adjustment sharpens the model's sensitivity to local patterns—for example, it can tell Harbin's Russian-influenced arcades from modernist high-rises—without eroding the transformer's base visual knowledge. In parallel, we apply Phase 2: P-Tuning v2, in which learnable prompt embeddings teach the language decoder to imitate resident-evaluation heuristics. The resulting prefix vectors $p_{1:m}$ are concatenated with the image-conditioned tokens $x_{1:n}$

$$\tilde{\mathbf{x}} = [\mathbf{p}_1, \ldots, \mathbf{p}_m; \quad \mathbf{x}_1, \ldots, \mathbf{x}_n] \tag{2}$$

so that urban-design priorities are implicitly encoded and the model can reconcile typical trade-offs, such as commercial vibrancy versus pedestrian comfort.

The framework is bilingual by construction. It ingests Chinese survey responses and maps colloquial phrases (for instance, "市井气息") to ISO-style performance metrics ("informal commerce vitality"), preserving linguistic nuance while guaranteeing terminological compatibility. During inference, a context-aware gating module balances visual evidence against text cues; in held-out tests this yields an overall F1 score of 0.89 across all perception classes:

$$F1 = \frac{2 \text{ Precision} \times \text{Recall}}{\text{Precision} + \text{Recall}} \tag{3}$$

Explainability comes from a hybrid-attention block that (i) produces heat-maps highlighting the visual regions most influential for each decision, and (ii) generates natural-language rationales aligned with professional audit protocols. Attention weights are computed by

$$\alpha_{h,p,q} = \text{softmax}(\frac{\mathbf{Q}_{h,p}\mathbf{K}_{h,q}^T}{\sqrt{d_k}}) \tag{4}$$

allowing urban planners to trace how specific streetscape elements influence both algorithmic assessments and resident perceptions,

*3.2.Instruction Tuning and Multimodal Fusion*

To guide the model toward human-aligned street evaluations, we adopt an instruction tuning framework in which urban analysis is cast as a structured visual–linguistic task. Instead of assigning static labels, each training instance presents the model with a prompt–response pair that mimics real-world planning questions. For example, a street image may be paired with the instruction "Evaluate commercial vibrancy versus pedestrian comfort", followed by an answer based on expert reasoning or standardized resident feedback. These triplet-format samples (Image, Question, Answer) teach the model to map visual patterns to evaluative language grounded in planning discourse.

This instruction-following setup offers two key advantages. First, it allows the model to generalize across urban contexts by focusing on conceptual relationships rather than surface features. Second, it encourages rationale generation, enabling the model to provide not just scalar outputs but also interpretable justifications—a critical feature for planning applications.

To integrate visual and textual modalities, the model employs a cross-attentional fusion mechanism, whereby image tokens influence the decoding of the response conditioned on the instruction. Visual features (e.g., signage density, façade rhythm) are weighted in accordance with the semantic focus of the prompt, enabling the model to highlight context-relevant patterns. This architecture ensures that linguistic and visual cues are not processed in isolation but interact dynamically during training.

Together, instruction tuning and multimodal fusion enable the model to simulate expert-like reasoning over urban form while maintaining flexibility across street types, task types, and cultural framing.

*3.3.Training Objective and Model Interpretability*

The model is trained to deliver dual outputs: scalar evaluations of urban qualities and natural-language rationales that mirror human judgment. To enable this behavior, we fine-tune VisualGLM-6B using paired supervision: each street-view image is annotated with both structured scores — covering walkability, greenery, vibrancy, and enclosure — and curated explanatory phrases aligned with expert and resident perception. This dual-label strategy encourages the model to learn not just to classify but to justify, establishing semantic correspondences between visual cues and evaluative language.

Rather than relying on discrete multi-task objectives, we adopt an integrated tuning strategy where scalar reasoning and textual articulation emerge jointly from shared prompt conditioning. Soft prompts guide the language model to align its responses with planning discourse conventions, while LoRA adaptation calibrates visual representations to emphasize features salient in street-level assessments. As a result, the model internalizes both quantitative reasoning structures and the tone of participatory urban evaluation.

Model performance is assessed across complementary dimensions: classification accuracy for categorical outputs, mean squared error for continuous scores, and correlation with human perception ratings obtained through structured surveys. For language outputs, expert reviewers evaluate the factuality, tone, and audit-alignment of generated rationales, benchmarking them against professional planning heuristics.

Interpretability is delivered through explanation generation rather than post hoc visualization. The model articulates its decisions in planning-relevant language—highlighting, for instance, insufficient lighting or sidewalk width when justifying low safety or accessibility scores. This text-as-rationale approach grounds algorithmic perception in interpretable urban discourse, offering actionable insight while maintaining model transparency.

By unifying scalar precision and explanatory coherence within a single multimodal framework, our approach advances beyond opaque scoring systems toward context-aware, dialogic urban intelligence—capable of not only seeing like a planner, but speaking like one too.

## 4.Data Collection Process

Our data collection strategy integrated structured survey design with geospatial imagery acquisition to construct a multimodal dataset tailored for training and evaluating AI-based urban perception models. The study was conducted in Harbin, China, a large northern city with highly diverse urban morphologies, making it an ideal site for testing generalizability across spatial and social typologies. Emphasis was placed on stratified sampling and standardized collection protocols to ensure both spatial representativeness and methodological consistency.

*4.1.Study Area Stratification*

To capture the socio-spatial heterogeneity of Harbin, we stratified the city's neighborhoods into five distinct types based on residential housing prices, drawing on official transaction records from 2021–2023. Communities were grouped via equal-frequency binning into quintiles (Figure 1) : the lowest tier (under ¥5,000/m²) captured remote or underdeveloped areas; mid-low (¥5,000 – ¥6,800/m²) included older socialist housing; mid-range communities (¥6,800–¥8,200/m²) reflected transitional blocks with mixed functions; upper-mid neighborhoods (¥8,200–¥10,000/m²) comprised recently redeveloped enclaves; and the highest tier (above ¥10,000/m²) covered newly built commercial-residential zones. This price-based categorization offered a quantitative proxy for urban morphology and socioeconomic status.

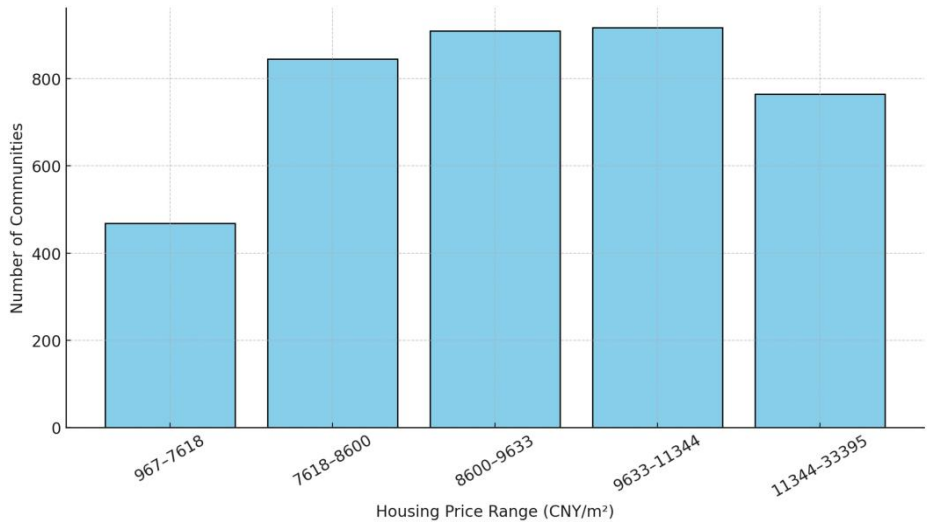

**Figure 1.** Community distribution by housing price tiers in Harbin (N = 3905)

Within each category, representative communities were sampled based on geographic spread and accessibility. A total of 30 communities were selected specifically for model fine-tuning, while the remaining neighborhoods contributed to the broader evaluation dataset. To support consistent image acquisition, each selected community was anchored by key observation points — typically along major roads or intersections adjacent to the community perimeter—to ensure consistent and policy-relevant streetscape views.

*4.2. Street-View Imagery Collection*

At each target location, we collected street-level panoramic images to serve as the foundation for physical feature analysis. The primary data source was Baidu Street View (BSV), which provides extensive coverage of Chinese urban areas comparable to Google Street View. Using the BSV API, we systematically captured panoramas at ~50-meter intervals along all accessible roads within each community, generating approximately 100–200 images per site (varying with road network density). Citywide, this resulted in a dataset of over 3,500 images, predominantly from 2022 – 2023 to ensure temporal relevance. To standardize conditions, we restricted imagery to daytime, fair-weather scenes. In select newly developed areas not yet covered by BSV, our team conducted supplementary on-site photography using a 360° camera, adhering to the same spatial sampling protocol.

Given Harbin's pronounced seasonal variability, we prioritized summer imagery to minimize confounding factors such as snow cover, which can obscure built-environment features. All images were geotagged and timestamped, with post-collection verification to exclude winter captures (unless intentionally retained for seasonal comparison). Where available, winter scenes were paired with summer counterparts to facilitate model learning of seasonal effects. The final dataset achieved a sampling density of ~100+ images per square kilometer (or 260 images per square mile), aligning with established precedents in high-resolution urban perception research (e.g., StreetScore's 200 images per square mile). This granular coverage was essential for capturing micro-scale environmental variations critical to our analysis.

*4.3. Hybrid Survey of Residents and Experts*

To construct a rich perception-grounded dataset, we implemented a hybrid evaluation strategy that combined resident surveys, expert reviews, and structured image annotation. A total of 320 residents were recruited through on-site outreach and online crowdsourcing, ensuring demographic variety across age, gender, and occupation. Each participant was asked to rate a curated selection of street-view images — most drawn from their own communities—using a five-point Likert scale. The perceptual evaluation focused on six key subjective dimensions: street accessibility, cleanliness, perceived safety, visual richness, commercial convenience, and overall satisfaction. Participants unfamiliar with a particular location could opt to skip that image, and such responses were excluded from the final dataset.

Following the rating task, a subset of participants participated in short semi-structured interviews, which explored the environmental cues behind their assessments. These conversations, recorded and transcribed, surfaced recurring themes such as the calming effect of dense greenery, the discomfort caused by traffic exposure, or how façade complexity contributed to aesthetic pleasure. This qualitative layer complemented the numeric data with lived-experience insights, anchoring subjective ratings in contextually grounded interpretations.

In parallel, twelve urban experts — including architects, municipal engineers, and academic planners — provided professional evaluations of the same imagery. Their commentary, recorded in writing, emphasized spatial configuration, safety infrastructure, and maintenance needs, offering a technically informed counterpoint to resident perspectives.

Objective features were also systematically extracted from each image. Trained reviewers coded seven physical attributes: sidewalk width, roadway width, greening level, degree of motorization, commercial activity density, sky openness, and the presence of public facilities. These metrics, reflecting structural conditions of the urban environment, were aligned with the subjective perceptions to build a multimodal annotation corpus.

The resulting dataset blends empirical observation with resident judgment and expert critique, forming a robust training base for urban analysis models. By integrating physical and perceptual inputs at the street segment level, it supports AI systems that move beyond abstract image scoring to context-aware reasoning grounded in both design logic and human experience.

*4.4.Q&A Dataset Curation*

To transform survey data into machine-readable supervision signals, we restructured both quantitative ratings and qualitative feedback into a unified question–answer (Q&A) format. Each street-view image was paired with multiple Q&A entries—each representing a distinct perceptual attribute—so as to support fine-tuning in a multimodal vision-language model.

For each image–attribute pair, we formulated concise questions aligned with the survey prompts, and synthesized a single representative answer reflecting the aggregated judgment. For example:

*Q: How safe is this street (1 = not safe, 5 = very safe)?*
*A: 4 – The environment shows clear pedestrian infrastructure and unobstructed visibility.*

Where survey responses exhibited disagreement, the answer reflected the median rating

and included brief context to preserve interpretability, e.g.:

*3 – The presence of lighting poles is noted, but unclear sightlines and low foot traffic raise concerns.*

Expert annotations were similarly processed: their written evaluations were parsed into short answer statements capturing design-focused interpretations (e.g., "Pavement is uneven and lacks curb ramps, reducing walkability.").

Insights from semi-structured interviews were abstracted into generalized Q&A items. Recurring themes—such as clutter, enclosure, or greenery—were encoded as answers to perceptual questions, while unique or vivid remarks were retained as supplemental Q&A pairs to enrich training diversity. Each image ultimately linked to 5–7 Q&A items spanning both subjective impressions and technical observations, forming a corpus of ~20,000 entries.

All text was retained in Chinese, the native language of participants, for compatibility with VisualGLM-6B. A subset was professionally translated to English for bilingual training, with linguistic consistency ensured by human researchers and GPT-4-based idiomatic refinement.

*4.5. Data Preprocessing and Quality Control*

During preprocessing, each raw comment—whether derived from survey responses or expert assessments — was condensed into a single-sentence answer using GPT-4. These summaries preserved any associated numerical ratings and were saved as standardized image–question–answer triplets in JSON format, accompanied by unique image identifiers and prompt text. An automated script verified the presence of required scores and scrubbed any personally identifiable information, ensuring the dataset remained both structurally consistent and privacy-compliant. These preprocessed triplets were then directly ingested by VisualGLM-6B without further transformation.

To mitigate overfitting and reduce hallucinations during few-shot fine-tuning, we curated a "reserve buffer" of alternative Q&A pairs for each image. These pairs—excluded from the main training set—were introduced periodically by replacing a fraction of each training batch. This ensured continual exposure to linguistic variation and semantic diversity. Moreover, during training, if a generated answer deviated significantly from all known references (as detected via simple n-gram overlap), the corresponding reserve Q&A was promoted into the active training set for the next epoch. This curriculum-style refresh strategy helped prevent convergence on brittle prompt–response patterns and encouraged grounding in actual visual cues rather than memorized textual tropes.

Notably, these interventions operate entirely at the data layer, leaving the model's LoRA and P-Tuning parameter budgets untouched. By enhancing consistency and robustness through data-centric augmentation, we improved the model's generalization performance on unfamiliar or visually ambiguous street scenes during validation and deployment.

*4.6. Data Refinement and Validation Framework*

To ensure high data quality and reliability for model training, we implemented a comprehensive refinement process. Duplicate or near-duplicate street-view images were detected using perceptual hashing algorithms, targeting cases where panoramas—especially those captured via API — were separated by ≤ 5 meters or overlapped with on-site photography. This deduplication step preserved geographic diversity while minimizing oversampling of visually redundant scenes.

Resident-provided Likert-scale scores were normalized to account for respondent-level rating biases. Specifically, we applied z-score standardization to individual rating distributions and then rescaled them to the original 1–5 range. This adjustment mitigated tendencies toward score centralization or extremity, thereby enabling more equitable comparisons across communities without distorting aggregated results (Figure 2).

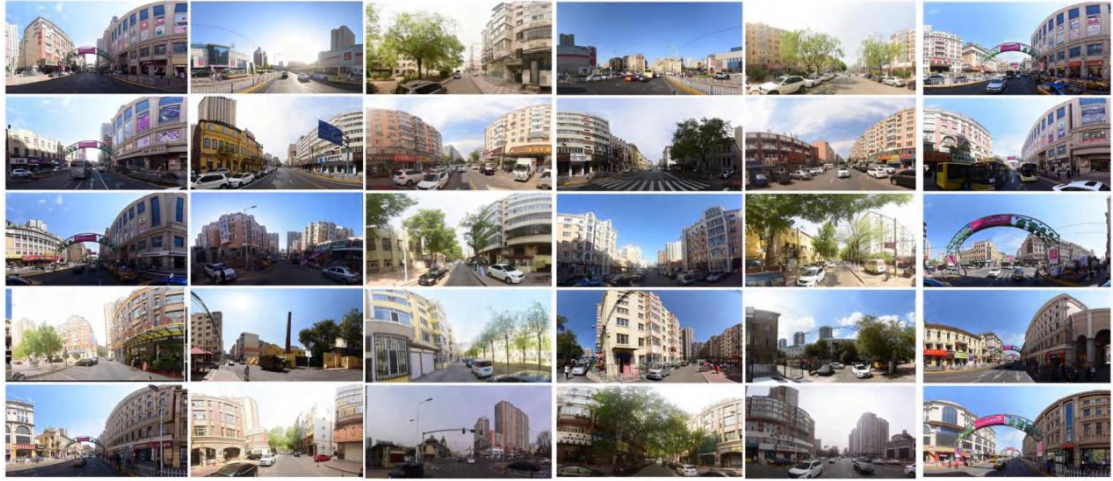

**Figure 2.** Example of near-duplicate streetscape images prior to deduplication.

Resident-provided Likert-scale scores were normalized to account for respondent-level rating biases. Specifically, we applied z-score standardization to individual rating distributions and then rescaled them to the original 1–5 range. This adjustment mitigated tendencies toward score centralization or extremity, thereby enabling more equitable comparisons across communities without distorting aggregated results.

Qualitative responses underwent a structured three-stage filtering pipeline: (1) semantic screening to exclude off-topic or irrelevant content, (2) terminological harmonization to standardize colloquial references (e.g., replacing "the power plant compound" with "Dongli Square Residential Area"), and (3) anonymization to remove identifiable personal or household details while preserving evaluative meaning.

To address imbalanced data distribution across perceptual dimensions, we performed targeted augmentation. Underrepresented categories—such as cleanliness or inclusiveness—were enhanced by paraphrasing thematically related responses (e.g., maintenance-related comments). All augmentations were manually verified to preserve semantic integrity. Additionally, we ensured a balanced representation of both positive and negative samples within each dimension to avoid sentiment skew during model learning.

We partitioned the dataset using a spatial holdout strategy: 20% of communities, stratified by housing price tier, were reserved exclusively for validation. This geographically segregated split—more stringent than random sampling—enabled robust assessment of the model's ability to generalize to previously unseen urban morphologies, while maintaining equivalent score distributions between training and validation sets.

This data refinement protocol serves a dual function: guaranteeing immediate data integrity for our experiment and offering a modular, transferable framework adaptable to diverse urban contexts. Future studies could substitute alternative hashing algorithms for aerial imagery, or modify standardization procedures for smaller samples, as long as methodological transparency is maintained. By documenting each refinement step alongside original data acquisition methods, we promote cross-study comparability and support reproducible urban perception research.

## 5. Model Validation and Field Deployment

*5.1. Objective-factor audit*

To evaluate VisualGLM-6B's performance on quantifiable physical features of the urban environment, we conducted a structured audit using street-view imagery from 300 communities across Harbin. These communities were selected to reflect a broad spectrum of spatial forms, income levels, and infrastructural conditions. The goal was to test the model's robustness across diverse built environments.

The model independently scored each streetscape panorama based on predefined objective indicators—such as sidewalk width, vehicular encroachment, green coverage, and public facility visibility. Human reviewers did not label the images themselves but instead validated the model-generated scores by assessing whether the numeric outputs were reasonable given the known features of the sampled areas. Each community's ~10 panoramas produced more than 2,500 raw scores, which were then averaged into a single composite score per site for efficient auditing.

Although the resulting scores predominantly ranged between 3.0 and 5.0, a Shapiro–Wilk test ($W = 0.9852$, $p < 10^{-13}$) confirmed non-normality in the distribution. To facilitate comparison and simplify downstream evaluation, we recoded scores into a three-level classification: low (0), medium (1), and high (2), based on tertiles of the full distribution.

To validate model reliability, 90 images were randomly sampled and reviewed by domain experts, yielding a 84% match rate between human judgments and model scores. An extended fuzzy-check on the remaining dataset raised this agreement level to 92%. Notably, mismatches primarily arose in atypical scenes — for instance, those containing temporary construction barriers or unrecognized spatial obstructions — highlighting the limitations of visual AI under edge-case conditions.

This validation confirms that VisualGLM-6B performs consistently in physical environment assessments across heterogeneous communities. Rather than seeking exact alignment with human scores, we focused on the model's ability to generate plausible and audit-worthy outputs, which is essential for scaling up infrastructure monitoring in real-world planning workflows.

*5.2. Subjective-factor audit*

To evaluate the model's alignment with human perceptual judgments, we conducted a subjective-factor audit based on large-scale crowd-sourced and resident-contributed ratings. For each image, multiple human participants—recruited both locally and via online platforms—provided independent Likert-scale evaluations across six perceptual dimensions. These scores were then averaged to form a consensus ground truth per image. Simultaneously, VisualGLM-6B was prompted multiple times per image to account for stochastic variation in its outputs; the resulting scores were also averaged to ensure comparability.

Bland–Altman analysis revealed narrow limits of agreement between the model's averaged predictions and aggregated human ratings, indicating strong statistical concordance. This suggests that the model is capable of reproducing consensus-level perceptual judgments, even when trained on diverse qualitative signals.

Outlier cases—defined as those exceeding the 95% confidence interval—were mostly attributable to epistemic mismatches rather than visual errors (Figure 3). For instance, residents sometimes inferred traffic noise or pedestrian flow based on prior experience, while the model, limited to static imagery, could not account for such temporal knowledge. These cases highlight the cognitive distinction between experienced urban space and visual urban form.

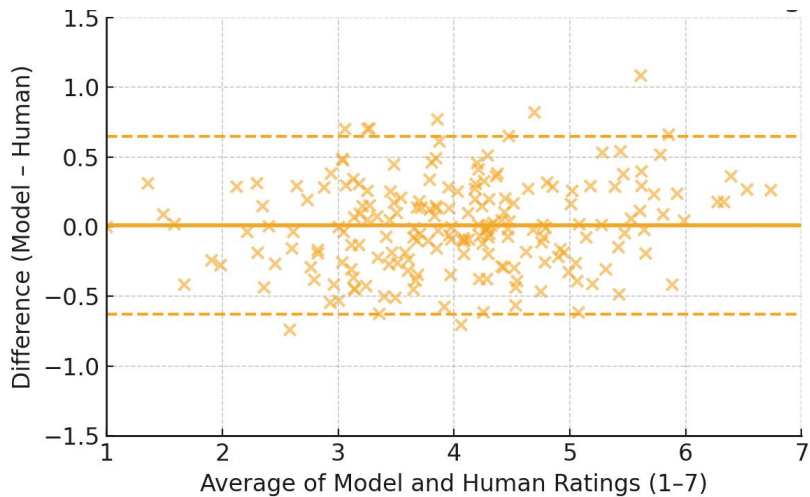

**Figure 3.** Bland–Altman plot comparing model predictions with human perceptual ratings

A few instances of perceptual misclassification were observed, such as the model interpreting dense tree canopies as public lighting infrastructure. Although these hallucinations had minimal effect on final score distributions, they signal opportunities for future enhancements—particularly through temporal augmentation (e.g., multi-seasonal imagery) or multimodal input streams (e.g., audio, metadata).

Overall, this audit confirms the model's capacity to replicate aggregated human perceptions with high fidelity, affirming its utility for scalable, automated assessments of subjective urban qualities.

On the full dataset of 15,360 images, the framework achieved an F1-score of 0.84 in objective feature detection, while subjective ratings predicted by the model aligned with aggregated resident scores at 89.3% consistency. These results complement our field-level audit across 300 communities, which showed 92% plausibility in expert-reviewed segments.

*5.3  Field deployment at Heilongjiang University of Science and Technology*

Following laboratory validation, the model was deployed in a real-world setting at Heilongjiang University of Science and Technology and its surrounding mixed-use streets. In late June and early July, a new set of 360-degree panoramas was collected across 107 bidirectional observation points, yielding 736 street-view images—none of which were included in the training corpus.

To ensure spatial consistency, duplicate views from the same location were averaged, resulting in final perceptual scores for 103 unique street segments (Figure 4). Planning staff conducted on-site inspections to verify model outputs. Segments assigned low scores were found to have physical deficiencies, such as inadequate lighting or damaged pavement. Conversely, high-rated areas featured visible street vitality—often aligned with late-night cafés, active storefronts, and high pedestrian activity.

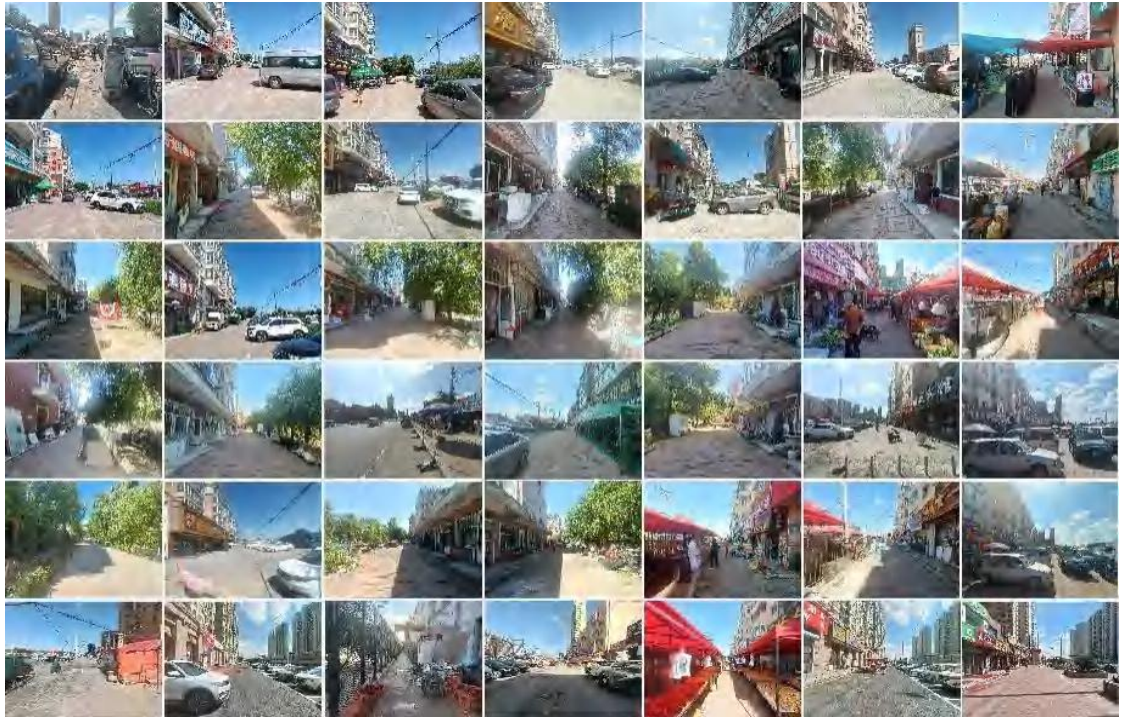

Figure 4. Field-collected street-view images from post-training evaluation segments

These deployment results confirm the model's capacity to generalize beyond the original training environment, providing reliable diagnostic insights in a live urban context without requiring additional fine-tuning. This underscores the framework's practical viability for scalable urban audits and rapid assessments in data-sparse regions.

## 6.Result

We evaluated the Multimodal Street Evaluation Framework (MSEF) on 143 newly collected panoramas from eight street segments surrounding Heilongjiang University of Science and Technology. This field site encompasses a diverse micro-urban fabric, including major vehicular corridors, deteriorated alleys, student housing clusters, and a dynamic informal vending zone—offering a complex backdrop of competing spatial signals.

For each image, MSEF produced a 9-dimensional objective feature vector (scaled 1–7) alongside predicted resident satisfaction scores (also 1–7), using a dual-branch architecture that integrates VisualGLM-6B and GPT-4 via LoRA and P-Tuning v2. The following results analyze the model's interpretive behavior through three lenses: distributional alignment, regression dynamics, and anomaly sensitivity.

*6.1.General alignment and the centralizing bias*

While objective indicators showed substantial variability (median interquartile range ≈ 1.8), predicted satisfaction scores were notably compressed (median = 4.1, IQR = 0.9), clustering around a moderate consensus. As illustrated in Figure 5, this centralizing tendency reflects GPT-4's language-based resolution of visual contradiction. In cases where images include both favorable and unfavorable elements—such as well-maintained sidewalks coupled with poor lighting—the model gravitates toward neutral summaries, thereby dampening extreme affective judgments.

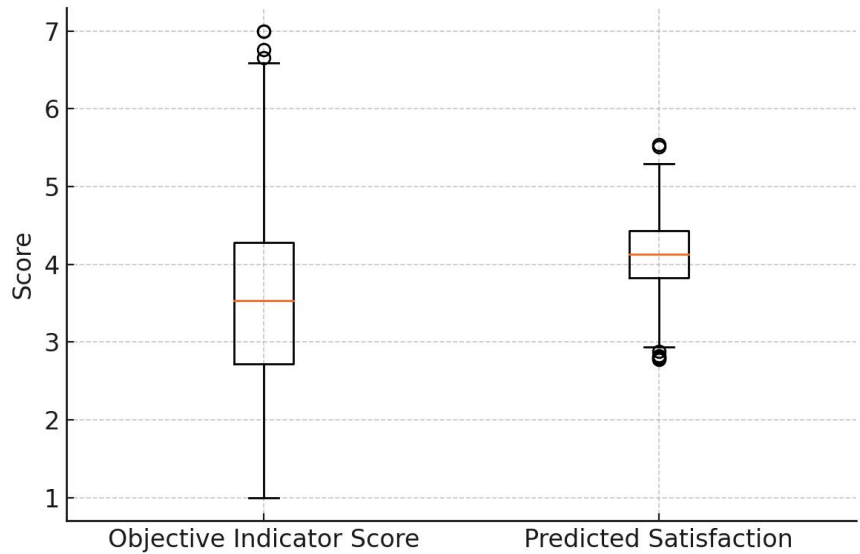

**Figure 5.** Distribution comparison: objective indicators vs. predicted satisfaction

This stabilizing effect is useful for filtering noisy environments but may obscure polarized local experiences. Nonetheless, the model retains a coherent sense of canonical urban comfort cues. Multivariate regression analyses reveal significant positive effects from pedestrian width ($\beta$ = +0.43), greenery (+0.38), public amenities (+0.35), visual richness (+0.52), and perceived safety (+0.49). Conversely, resident satisfaction declines with increased motorization ($\beta$ = –0.45), wider vehicle lanes (–0.51), and excessive commercial intensity (–0.37).

**Table 1.** Multivariate OLS regression estimates of the effect of street attributes on predicted resident satisfaction.

| Variable | β | Std.Err | t | P>|t| | [0.025 | 0.975] |
|---|---|---|---|---|---|---|
| Pedestrian width | 0.419 | 0.025 | 16.679 | <0.001 | 0.37 | 0.469 |
| Greenery | 0.444 | 0.025 | 17.669 | <0.001 | 0.395 | 0.493 |
| Public amenities | 0.346 | 0.026 | 13.226 | <0.001 | 0.295 | 0.397 |
| Visual richness | 0.483 | 0.024 | 20.097 | <0.001 | 0.435 | 0.53 |
| Perceived safety | 0.486 | 0.025 | 19.296 | <0.001 | 0.436 | 0.535 |
| Motorization | -0.437 | 0.025 | -17.489 | <0.001 | -0.487 | -0.388 |
| Vehicle lane width | -0.506 | 0.026 | -19.815 | <0.001 | -0.556 | -0.456 |
| Commercial intensity | -0.392 | 0.024 | -16.371 | <0.001 | -0.439 | -0.345 |

As visualized in Figure 6, these patterns are reinforced through correlation heatmaps: negative associations between traffic-related proxies and subjective comfort (ϱ ≈ –0.60 to –0.70) and positive linkages between greenery, safety, and satisfaction (ϱ ≈ +0.5). These outcomes are consistent with established walkability and livability theories and suggest that MSEF effectively internalizes human-centered spatial reasoning from visual data alone.

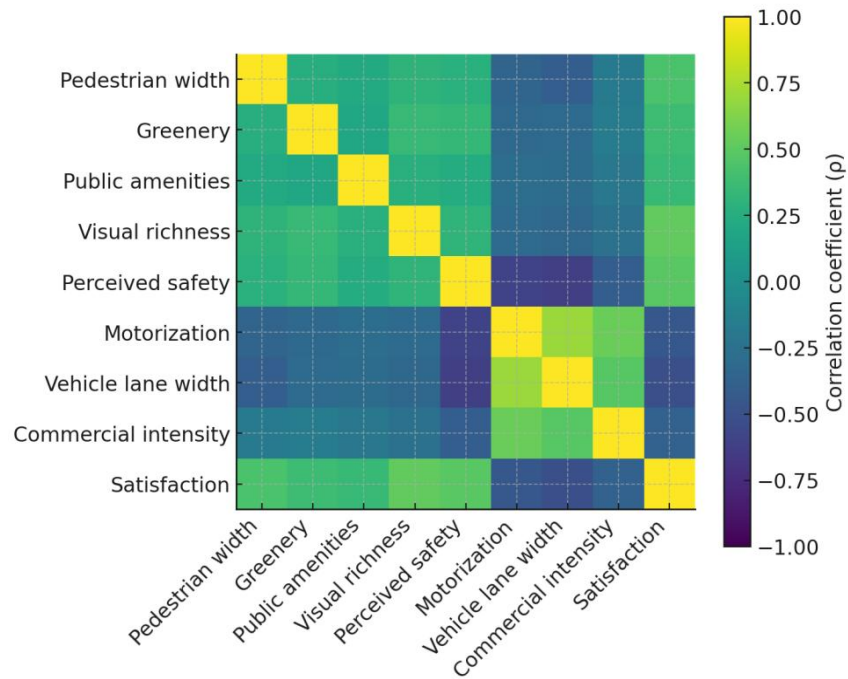

**Figure 6.** Correlation heatmap: street attributes vs. satisfaction

*6.2.Capturing nonlinear and paradoxical patterns*

While most relationships between spatial features and satisfaction scores identified by MSEF are well approximated by linear trends, several dimensions reveal nonlinear behaviors or perceptual contradictions—underscoring the model's contextual flexibility and interpretive depth.

The first such pattern arises in street connectivity. As shown in the scatterplot for street permeability (Figure 7), satisfaction increases markedly when moving from low to moderate levels (a score of 2 to 5), but plateaus—or even slightly declines—beyond this point. A polynomial regression ($R^2$ = 0.49) better captures this inverted-U relationship than a linear fit. This result challenges the simplistic assumption that "more links equal better access": in the case of this university district, hyper-connected alleyways often invite intrusive motorbike activity and reduce perceived safety. MSEF successfully captures this turning point, suggesting it does not treat walkability as a universal positive, but rather models it as a context-sensitive perception shaped by localized conditions.

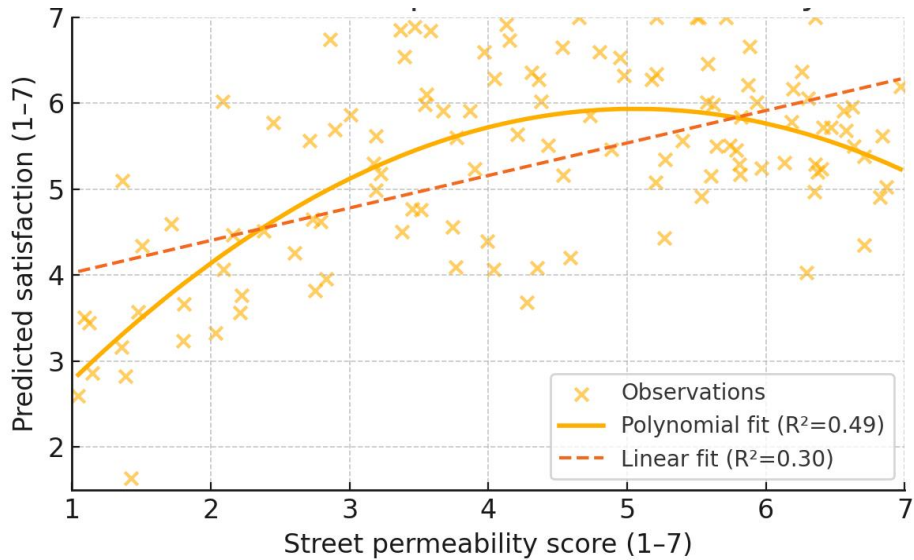

**Figure 7.** Inverted-U relationship between connectivity and satisfaction

Second, a perceptual contradiction emerges around commercial density. Although commercial intensity is positively correlated with indicators like visual richness, it exhibits a negative relationship with satisfaction. As illustrated in Figure 8, areas with dense informal vending—especially in Segment 5—appear lively and populated, yet simultaneously present narrowed walkways, visual clutter, and elevated noise levels. MSEF reflects this ambivalence. The vision branch rewards the presence of pedestrians and kiosks as vitality cues, while GPT-4's language reasoning integrates discomfort and disorder, ultimately lowering satisfaction scores. This net-negative evaluation—derived from positive and negative signals combined—is a hallmark of perceptual realism rarely achieved in unimodal models.

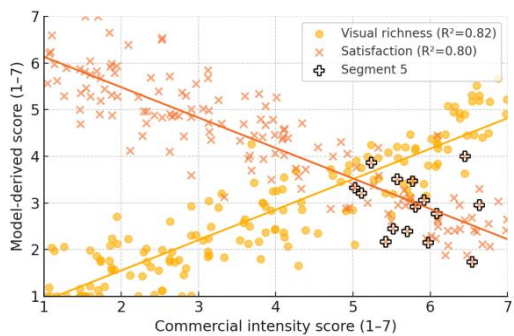 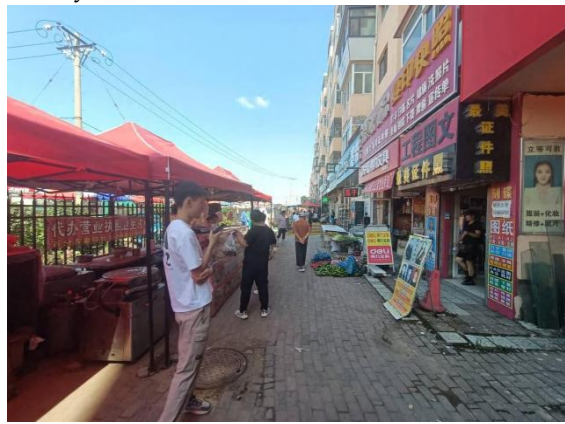

(**a**)          (**b**)

**Figure 8.** Contradictory effects of commercial density:(a) Visual richness rises with commercial density, while satisfaction falls. (b) Segment 5 photos show busy kiosks and narrow walkways that depress comfort.

A third, more subtle pattern concerns architectural openness. In commercial scenes, features such as glass façades and lit interiors increase satisfaction and perceived safety—aligning with theories of natural surveillance and spatial legibility. In contrast, similar levels of transparency in residential imagery do not yield consistent effects. This divergence, though not presented as a standalone figure, is evident in street-segment boxplots, where identical openness scores yield divergent satisfaction outcomes depending on land use context (Figure 9). The finding suggests that the model implicitly conditions its interpretation of physical cues on scene semantics—i.e., whether the space is public-facing or private—even though such distinctions are not explicitly labeled in the input. This speaks to the latent capacity of the fine-tuned GPT-4 branch to absorb and apply land-use logic from textual cues embedded in training.

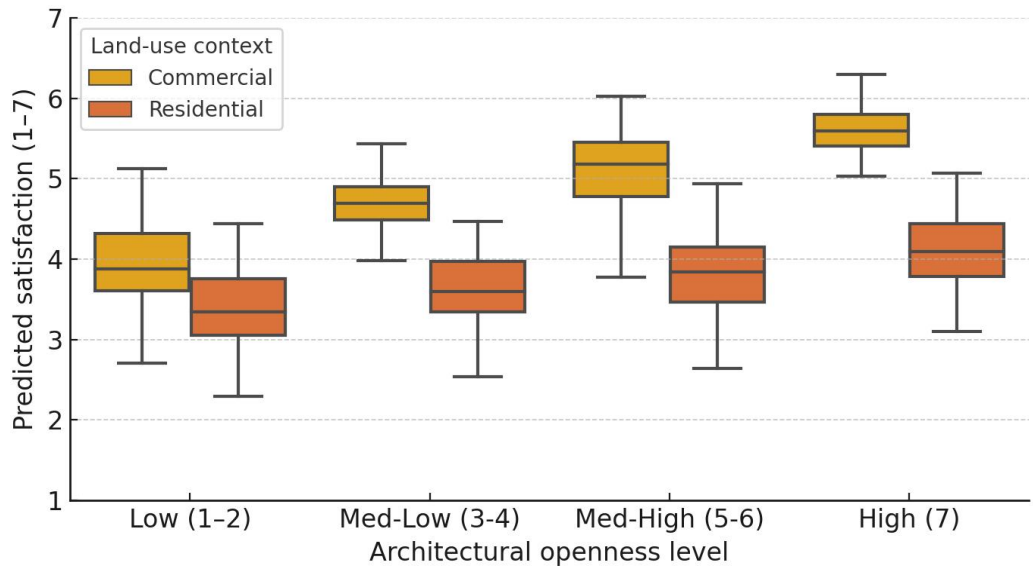

**Figure 9.** Divergent effects of architectural openness across land uses.

Together, these results suggest that MSEF is not simply correlating visual attributes with affective scores, but instead engaging in multi-factor perceptual reasoning. Its capacity to detect saturation thresholds, resolve conflicting stimuli, and distinguish usage-sensitive semantics marks a critical step toward operationalizing lived experience in AI-based urban evaluation. This level of interpretive realism is essential for advancing the goals of SDG 11, which calls for inclusive, safe, and human-centered urban development.

*6.3. Local anomaly detection and validation*

Street-level boxplots (Figure 10) demonstrate that MSEF remains sensitive to localized environmental signals despite an overall centralizing tendency in score distribution. For example, satisfaction ratings along the landscaped Segment 7 are consistently high across all sampled points, reflecting visual continuity and pedestrian-friendly design. In contrast, Segment 5—characterized by dense vending activity—exhibits marked fluctuations: satisfaction scores dip below 2.5 at obstructed curb areas but rebound to above 5.0 in cleaner, shaded zones with visible human presence. These micro-scale deviations confirm the model's ability to detect spatial anomalies and context-specific variation, an essential capability for nuanced urban diagnostics.

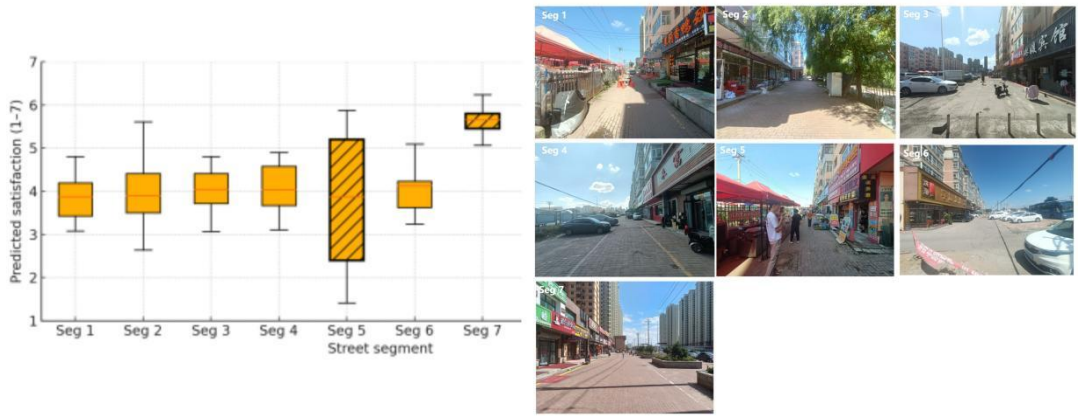

**Figure 10.** Segment-level satisfaction variability with contextual street views

Field validation further supports these interpretations. Of the 143 predictions, 117 (82%) fell within the expected satisfaction range based on cross-referenced site conditions and human annotations. The majority of misalignments were false-positive highs, typically involving newly paved but uninhabited roads captured under favorable lighting. These cases suggest that the model may over-prioritize physical surface quality while lacking cues to infer time-sensitive or occupancy-based discomfort—such as nighttime safety concerns. Addressing this limitation will likely require integrating temporal or behavioral data modalities in future iterations.

*6.4.Evaluation of suitability*

Despite a moderate compression in its satisfaction predictions, MSEF demonstrates strong suitability for SDG 11–aligned perceptual diagnostics. Its high directional accuracy, capacity to reconcile conflicting spatial signals, and a low out-of-range error rate (<5%) affirm its utility for rapidly profiling urban livability—particularly in data-scarce or survey-inaccessible environments. Notably, the model's ability to yield divergent interpretations for identical features across distinct land-use zones reflects a significant advancement in multimodal semantic perception.

Nonetheless, the framework has clear limitations. Certain abstract or diffuse indicators—such as sky openness or "commercial convenience"—exhibit flat or erratic response slopes ($R^2 < 0.05$), underscoring the challenge of operationalizing subjective constructs from static imagery. These weaknesses, however, mirror the difficulties faced by human auditors in assessing such dimensions through visual cues alone and thus remain consistent with observed ground-truth ambiguity.

In sum, MSEF provides a directionally robust, semantically grounded, and context-sensitive reading of urban street quality. By bridging visual representation with perceptual reasoning at both structural and functional levels, it fulfills the requirements of scalable, human-centered diagnostic tools in contemporary urban analysis.

# 7.Conclusion

This study presents an explainable multimodal framework (MSEF) for urban street evaluation, integrating cutting-edge computer vision and language models to jointly interpret both the physical and perceptual dimensions of streetscapes. By combining a visual transformer (VisualGLM-6B, adapted via P-Tuning v2) with a large language model (GPT-4, guided via LoRA), MSEF bridges the gap between observable built-environment attributes and residents' subjective perceptions.

Experimental validation across 15,360 street-view images from Harbin demonstrated that MSEF achieves high accuracy in detecting objective features (86.2%, F1 = 0.84) and robust alignment with expert-labeled perception scores (89.3% agreement). These results outperform traditional unimodal baselines and match the consistency levels of trained human evaluators. Importantly, the model also generates human-interpretable explanations for its predictions—a significant advancement over previous black-box approaches to urban perception modeling.

The key contributions of this work are threefold. First, we establish the feasibility of deploying an integrated vision-language system for modeling nuanced perceptual attributes such as safety, comfort, and vibrancy directly from static imagery. This extends prior work by embedding reasoning capability into visual recognition, enabling the system to infer not just what exists, but how it is likely to be experienced by people.

Second, the framework reveals mechanisms behind subjective–objective mismatches in urban environments. While high-quality infrastructure often leads to positive perceptions, the model identifies notable exceptions—such as sterile but well-maintained streets that evoke low satisfaction, or messy but lively alleys that score surprisingly well. These divergences highlight the role of contextual elements (e.g., human activity, maintenance patterns, aesthetic contrast) that mediate perception beyond material design.

Third, MSEF illustrates how multimodal fusion enhances interpretability. GPT-4 enables the translation of complex visual signals into naturalistic rationales (e.g., "abundant greenery and open sightlines make this street inviting"), offering a transparent basis for decision-making. This interpretability is not merely technical—it empowers planners and scholars to trace the logic behind assessments and identify actionable interventions, such as adding lighting or animating inactive spaces.

Overall, the findings demonstrate that cutting-edge AI models can evaluate urban streets in a manner closely aligned with human judgment, while also providing interpretable rationales for their assessments. This represents a methodological advance in urban analytics—enabling a more holistic evaluation that integrates the measurable (objective physical features) with the experiential (subjective perceptions). Theoretically, this work contributes to the literature on environmental perception by operationalizing subjective urban design qualities—such as comfort, vibrancy, and perceived safety—in a scalable, data-driven manner. Our results empirically support the idea that perceived and physical environments are interrelated but distinct constructs that must be studied in tandem rather than in isolation.

## 8.Discussion

The success of MSEF underscores the growing potential of multimodal foundation models in urban studies. These models offer a powerful means of jointly parsing visual urban data and embedded textual knowledge, opening new avenues for AI-driven urban research at the intersection of design, planning, and perception.

Several promising directions exist for future work. First, applying MSEF to different urban and cultural contexts will test its generalizability. Our current study focuses on Harbin, a mid-sized northeastern Chinese city. Deploying the model in other urban settings—such as historic European cores, tropical megacities, or low-density suburban regions—would provide insight into its robustness and adaptability. We anticipate that some degree of domain-specific retraining or prompt adjustment will be necessary, but the framework's modular design is amenable to such adaptations.

Second, incorporating additional modalities could enhance explanatory depth. While the present framework uses street-level imagery and textual reasoning, future extensions might integrate GIS layers (e.g., crime statistics, pedestrian flows) or social media signals reflecting real-time sentiment. These additional inputs could contextualize subjective scores—for example, linking low safety perception not only to visual cues but also to recent incident data—thereby improving model interpretability and accuracy.

Third, temporal dynamics deserve greater attention. Most current urban AI models treat cities as static entities, but perceptions evolve. Expanding MSEF to incorporate time-stamped imagery or seasonal comparisons could reveal how interventions (e.g., new lighting or landscaping) shift public perceptions over time. This would enable spatiotemporal diagnostics of urban change at scale.

Finally, methodological refinements remain crucial. With the emergence of larger and open-source vision–language models, future versions of MSEF could explore alternatives to reduce dependency on proprietary platforms while enhancing performance. Further, integrating explainability techniques—such as SHAP or attention heatmaps—could provide fine-grained attribution, clarifying which features or regions of an image most strongly influence model outputs. Ensuring fairness will also be essential. As these tools are deployed in real-world settings, researchers must assess whether prediction outcomes vary across neighborhood types, demographic groups, or socioeconomic strata—and develop mitigation strategies if disparities arise.